\documentclass[lettersize,journal]{IEEEtran}
\usepackage{amsmath,amsfonts}
\usepackage{algorithmic}
\usepackage{algorithm}
\usepackage{array}
\usepackage{float}
\usepackage[caption=false,font=footnotesize,labelfont=rm,textfont=rm]{subfig}
\usepackage{textcomp}
\usepackage{stfloats}
\usepackage{url}
\usepackage{verbatim}
\usepackage{cite}
\usepackage{tabularx}
\usepackage{makecell}
\usepackage{romannum}
\usepackage{graphicx}

\begin{document}
\bstctlcite{IEEEexample:BSTcontrol}

\title{Large Language Model guided Deep Reinforcement Learning for Decision Making in Autonomous Driving}
\author{Hao Pang, Zhenpo Wang, Guoqiang Li
\thanks{The authors are with the School of Mechanical Engineering, Beijing Institute of Technology, Beijing 100081, China (e-mail:
bitpang@bit.edu.com; wangzhenpo@bit.edu.cn; guoqiangli@bit.edu.cn).} }

\maketitle

\begin{abstract}
Deep reinforcement learning (DRL) shows promising potential for autonomous driving decision-making. However, DRL demands extensive computational resources to achieve a qualified policy in complex driving scenarios due to its low learning efficiency. Moreover, leveraging expert guidance from human to enhance DRL performance incurs prohibitively high labor costs, which limits its practical application. In this study, we propose a novel large language model (LLM) guided deep reinforcement learning (LGDRL) framework for addressing the decision-making problem of autonomous vehicles. Within this framework, an LLM-based driving expert is integrated into the DRL to provide intelligent guidance for the learning process of DRL. Subsequently, in order to efficiently utilize the guidance of the LLM expert to enhance the performance of DRL decision-making policies, the learning and interaction process of DRL is enhanced through an innovative expert policy constrained algorithm and a novel LLM-intervened interaction mechanism. Experimental results demonstrate that our method not only achieves superior driving performance with a 90\% task success rate but also significantly improves the learning efficiency and expert guidance utilization efficiency compared to state-of-the-art baseline algorithms. Moreover, the proposed method enables the DRL agent to maintain consistent and reliable performance in the absence of LLM expert guidance. The code and supplementary videos are available at https://bitmobility.github.io/LGDRL/.
\end{abstract}


\section{Introduction}
\IEEEPARstart {I}{n} recent years, autonomous driving technology has gained significant attention and extensive research due to its immense potential to revolutionize the automotive industry \cite{ir1}. Autonomous vehicles must navigate complex traffic scenarios with multiple dynamic agents, necessitating sophisticated decision-making capabilities. The behavior decision-making system represents a critical component of autonomous vehicles, responsible for generating safe and reasonable driving behaviors\cite{decisionmaking}. However, existing decision-making approaches confront substantial challenges in handling complex interactive driving scenarios. The rule-based methods demonstrate inherent limitations in generating adaptive and context-aware decisions in driving scenarios involving intricate multi-agent interactions\cite{rulebased}. As one of the most advanced artificial intelligence (AI) technologies, deep reinforcement learning (DRL) has achieved remarkable success in addressing a series of challenging decision-making tasks \cite{doudizhu, game2}. Applying DRL to develop highly intelligent and reliable behavior decision-making systems for autonomous vehicles has become a focal point for researchers\cite{ir5}.

\begin{figure}[t]
    \centering
    \captionsetup{justification=justified, font={small,rm}} 
    \subfloat[In complex driving scenarios, traditional DRL approaches struggle to acquire sufficient successful trajectories, leading to low learning efficiency and impeding the development of qualified decision-making policies.]{\includegraphics[width=0.99\linewidth]{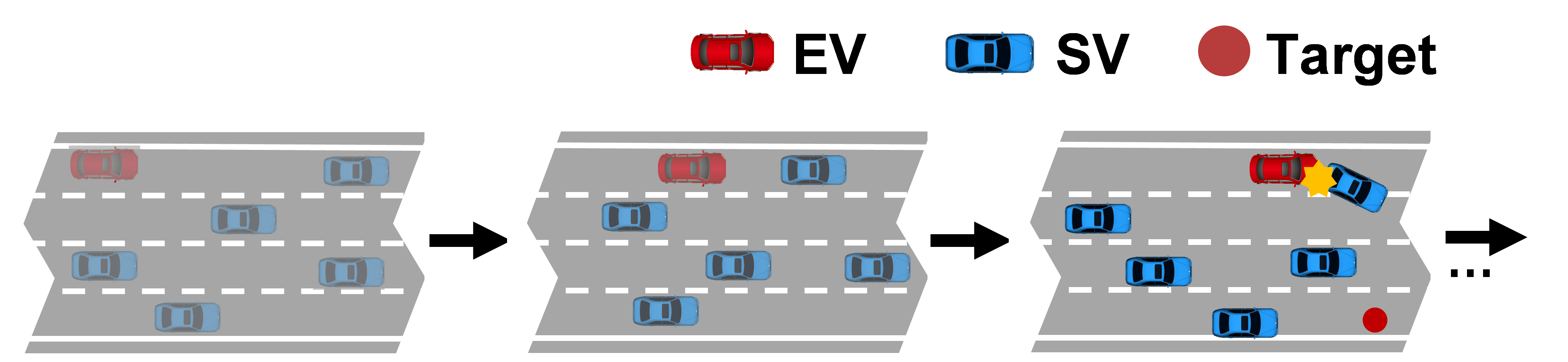}}
    \vspace{.04in}
    \subfloat[Our proposed LGDRL framework incorporates an LLM expert, which is prompted by a prompt generator, to provide guidance throughout the DRL learning process. It effectively enhances DRL learning efficiency and the quality of learned decision-making policies compared to traditional DRL methodologies.]{\includegraphics[width=0.99\linewidth]{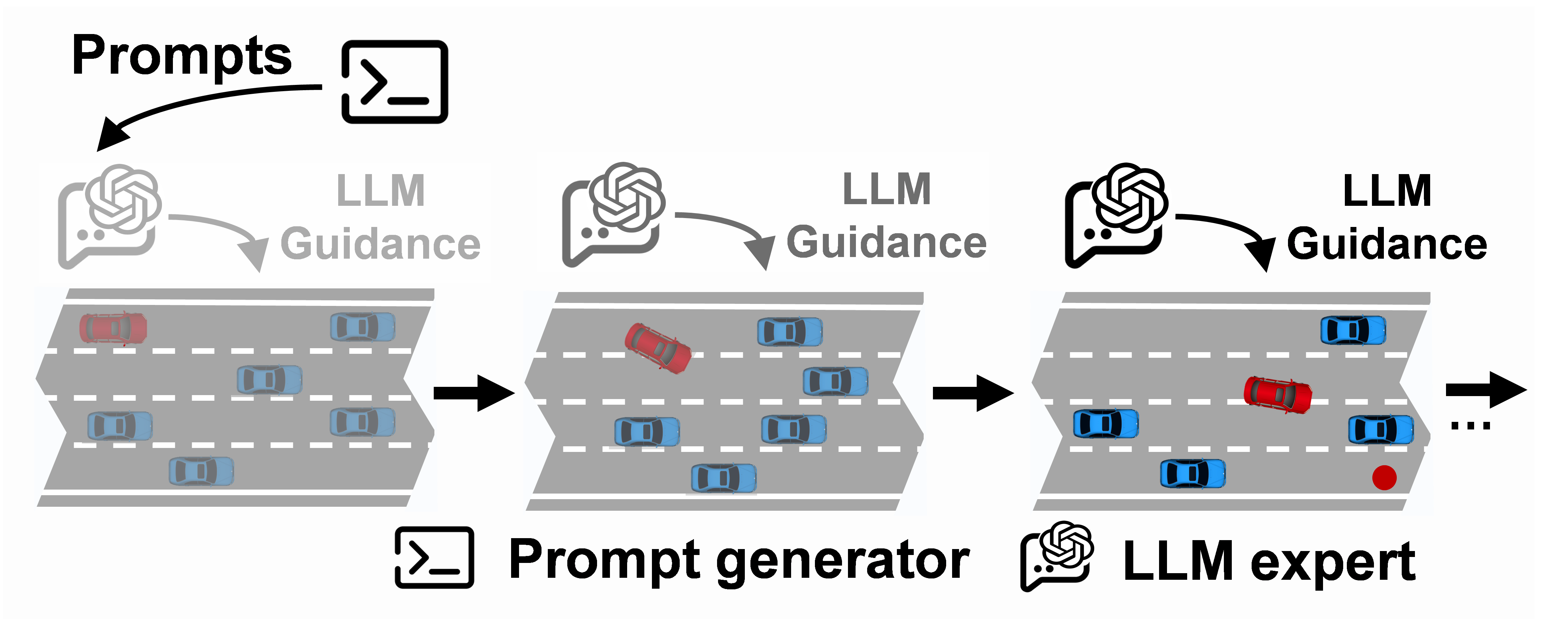}}
    
    \caption{Comparison of traditional DRL and the proposed LGDRL framework.}
    \label{fig:introduction}
\end{figure}

The application of DRL in behavior decision-making systems remains challenging, primarily due to its inherent low learning efficiency. DRL optimizes its policy through the sample data derived from the interactions with the environment. As illustrated in Fig. \ref{fig:introduction}(a), traditional DRL approaches struggle to accumulate sufficient successful trajectories in complex driving scenarios, leading to low learning efficiency. This limitation critically impedes the policy optimization process of DRL and often results in unqualified decision-making policies. Such a slow and unstable training process significantly hinders the practical deployment of DRL-based behavior decision-making systems in autonomous vehicles. \cite{ir8}.

Incorporating expert guidance into the learning process presents a promising solution\cite{irhumanguide}. However, existing methods typically demand substantial and continuous expert-level guidance from human drivers to satisfy the quality and quantity requirements for improving DRL-based decision-making systems, which results in high labor and time costs \cite{humanAi}. Furthermore, the inefficient utilization of expert guidance making it challenging for DRL agents to fully absorb and leverage expert guidance during the learning process \cite{expertai}. This limitation further restricts the performance of DRL-based decision-making systems. 

Facing these problems, this article propose a novel \textbf{L}LM \textbf{G}uided \textbf{D}eep \textbf{R}einforcement \textbf{L}earning (\textbf{LGDRL}) framework. As shown in Fig. \ref{fig:introduction}(b),  we leverage an LLM-based driving expert to provide guidance for enhancing the DRL learning process. The LLM-based driving expert eliminates the necessity for human expert guidance in the numerous interactions between the DRL agent and the environment. Subsequently, we introduce an innovative expert policy constrained DRL algorithm, which incorporates a policy constraint based on Jensen-Shannon (JS) divergence into the actor-critic framework. The policy constraint modifies the learning objective of DRL by limiting the divergence between the DRL policy and the LLM expert policy within a predefined boundary. Furthermore, the interaction between the DRL agent and the environment during training is modified by a novel LLM-intervened interaction mechanism, which enables the LLM expert to intermittently intervene in the interaction between the DRL agent and the environment. The main contributions of this study are summarized as follows:

1) We propose a novel LGDRL framework for the decision-making problem in autonomous driving. This framework integrates the LLM expert into the DRL training loop. The LLM expert is used to provide guidance for the DRL learning process, which significantly improves the learning efficiency and the driving performance of the DRL decision-making policy while also reducing the workload of humans.

2) To enhance the learning process of DRL, we introduce an innovative policy-constrained algorithm that effectively leverages guidance from the LLM expert. Compared to the baseline algorithms, our algorithm demonstrates highest learning efficiency and superior performance in terms of driving safety and task success rate.

3) We propose a novel LLM-intervened interaction mechanism to improve the DRL agent's decision-making capabilities through intermittent interventions. Experimental results show that the proposed mechanism preserves the agent's self-exploration capabilities, ensuring consistent and reliable performance in the absence of LLM expert guidance.

The structure of this article is organized as follows: Section \Romannum{2} reviews the literature related to this study. Section \Romannum{3} presents the problem formulation. Section \Romannum{4} elaborates on the proposed LGDRL framework. Section \Romannum{5} describes the experimental setup. The experimental results and discussion are presented in the subsequent Section \Romannum{6}. Finally, Section \Romannum{7} summarizes the main conclusions of our research.

\section{Related Work}
\subsection{DRL-Based Decision-making for Autonomous vehicles}
DRL has become pivotal in the development of autonomous driving decision-making systems. Its ability to tackle complex, high-dimensional issues significantly boosts the intelligence of these systems \cite{rrlc1}. Considering the varying driving styles of other vehicles, the proximal policy optimization algorithm was enhanced with a self-attention mechanism to tackle decision-making challenges. This integration resulted in more human-like behavior for autonomous vehicles \cite{rrlc2}. To ensure driving safety, a risk-aware decision-making DRL method was proposed to find a strategy with the minimum expected risk \cite{rrlc3}. In another study, a soft actor-critic (SAC) algorithm was used as a decision-making controller for continuous action spaces in highway driving scenarios \cite{rrlc4}. Additionally, the challenge of merging onto on-ramps was framed as a decentralized multi-agent reinforcement learning problem, where an action masking scheme was implemented to enhance learning efficiency by eliminating unsafe actions \cite{rrlc5}. Despite the progress made by these DRL-based approaches in autonomous vehicle decision-making, they primarily rely on trial and error interactions with the environment for policy optimization. This poses a challenge in complex driving scenarios, where acquiring high-quality training data is difficult, leading to low sample efficiency and learning efficiency. To overcome these limitations, integrating expert prior knowledge into DRL emerges a promising solution. This can be achieved through various means, such as expert evaluation, demonstration, and intervention.

Expert evaluation involves experts providing qualitative or quantitative feedback on the agent's behavior. In the context of behavior planning for autonomous vehicles, human rating systems were employed to guide DRL-driven unmanned ground vehicles \cite{rr3}. Additionally, ranking the decision trajectories of agents by human experts could provide qualitative assessments, aligning agent policies with human preferences \cite{rr4}. However, explicit evaluation is often impractical in complex scenarios. Expert intervention directly intervene in the agent's training process to prevent catastrophic behaviors. This approach is particularly useful in high-risk environments such as autonomous driving. Human intervention was integrated into autonomous vehicles, which terminated training tasks when necessary to prevent the vehicle from entering dangerous states \cite{rr6}. A physics-based model was integrated with human oversight to guide DRL training. This hybrid approach enhanced safety, even under conditions of degraded human feedback, establishing a robust performance baseline for the DRL agents\cite{rr7}. However, this kind of supervision requires real-time responses from humans during training. Expert demonstration leverages previously collected data to train DRL agents. Driving demonstration data from human experts was pre-stored in the replay buffer before training to enhance the safety of the DRL decision-making agent in complex lane-changing tasks \cite{rr9}. 

Although combining expert prior knowledge with DRL offers significant benefits, several challenges persist. Most methods rely on human experts to guide the learning process. However, this approach involves high labor and time costs. Therefore, it is essential to explore advanced AI technologies to provide high-quality prior knowledge for DRL, reducing the reliance on human participants.

\subsection{LLM for Autonomous Driving}
LLMs demonstrate extraordinary performance in the field of text generation and have been applied across various domains due to their remarkable reasoning capabilities and generalization potential \cite{ir14,ir15}. Their application in autonomous driving has also been explored \cite{ir16,ir17}. In the context of autonomous driving systems, LLMs can be leveraged primarily through two approaches: fine-tuning pre-trained models and prompt engineering \cite{ir18, ir19}.

In the fine-tuning approach, pretrained LLMs are adapted to specific autonomous driving tasks through additional training. GPT-Driver reframed motion planning as a language modeling problem, applying the OpenAI GPT-3.5 model directly to autonomous driving motion planning tasks through a prompt-reasoning-fine-tuning strategy \cite{rr10}. Similarly, MTD-GPT transformed multi-task decision-making problems into sequence modeling tasks, addressing various decision-making scenarios at unsignalized intersections and demonstrating robust generalization performance \cite{rr11}. LMDrive introduced a multimodal LLM autonomous driving agent to predict vehicle control signals \cite{rr18}. DriveGPT4 proposed a multimodal LLM-based autonomous driving agent that could not only predict basic vehicle control signals but also interpret the rationale behind actions in real time \cite{rr19}. Co-pilot utilized an LLM to complete specific driving tasks that align with human intentions based on provided information \cite{rr21}. However, its validation in complex interaction scenarios remained to be addressed.

Prompt engineering, on the other hand, utilizes the reasoning capabilities of LLMs by designing specific prompts without additional training. DiLu and Drive as You Speak employed LLMs as reasoning agents for closed-loop driving tasks, integrating memory modules to record experiences and using LLMs for reasoning and reflection \cite{rr12, rr13}. Additionally, Drive Like a Human used LLMs to understand driving environments in a human-like manner, constructing a closed-loop system that showcased the LLM's understanding and interaction capabilities in driving scenarios \cite{rr14}. LanguageMPC designed a Chain-of-Thought (CoT) framework for LLM inference, integrating low-level controllers with parameter matrices guided by the LLM \cite{rr22}. This approach outperformed traditional MPC and RL-based methods in simplified environments. However, its performance in complex scenarios was not been validated. Talk2Drive developed an LLM-based autonomous driving agent that made driving decisions based on contextual information and human language instructions, enabling personalized driving styles and preferences \cite{rr23}.
computational complexity

Despite the promising performance of these methods, significant challenges persist in their practical application. The computational complexity and the temporal inefficiency of LLM reasoning fundamentally conflicts with the real-time decision-making requirements of autonomous vehicles. Additionally, the substantial computational and memory demands of these models introduce critical deployment constraints in resource-limited vehicular systems. Moreover, the absence of explicit policy optimization frameworks prevents LLMs from systematically improving driving strategies through iterative learning.

\section{Problem Formulation}
We first formulate the behavior decision-making problem in autonomous driving as a Markov decision process within the framework of DRL. Then, the main components of the DRL-based decision-making policy, including state space, action space, reward function are elaborated in detail.

\subsection{Formulation of Behavior Decision-Making Problem}
For the behavior decision-making problem of autonomous vehicles, the DRL-based decision-making agent generates driving behavior at each time step based on current environmental state. Subsequently, the autonomous vehicle executes control actions based on the driving behavior to transition to a new state, while the DRL agent receives a corresponding reward signal based on updated state. This entire decision-making process can be formulated as a Markov decision process (MDP), which can be characterized by a tuple:
\begin{equation}
MDP = (\mathcal{S}, \mathcal{A}, r, p, \gamma),
\end{equation} 
Here, \(\mathcal{S} \in \mathbb{R}^n\) represents the state space. The action space for the DRL agent is denoted by \(\mathcal{A} \in \mathbb{R}^n\). The reward function is defined as \(r: \mathcal{S} \times \mathcal{A} \to \mathbb{R}\). The state transition probability \(p: \mathcal{S} \times \mathcal{A} \times \mathcal{S} \to [0, 1]\) indicates the probability distribution of transitioning to the next state \(s' \in \mathcal{S}\) given the current state \(s \in \mathcal{S}\) and action \(a \in \mathcal{A}\). The discount factor \(\gamma \in (0, 1]\) adjusts the importance of future rewards. The learning objective of DRL is defined as follows:
\begin{equation}\label{eq:obj_1}
\begin{aligned}
&\mathop{\max}_{\pi} J(\pi) =  \mathop{\max}_{\pi} \mathbb{E}_{(s_t,a_t) \sim \rho_{\pi}}\left[\sum_{t=0}^{T} \gamma^t r(s_t, a_t)\right],
\end{aligned}
\end{equation}
where  \(\pi\) represent the DRL policy. The trajectory distribution induced by the DRL policy is denoted as \(\rho_{\pi}\). \(T\) represents the time horizon.

\subsection{Formulation of DRL-based Decision-Making Policy}

1) State space. The state space is composed of the states of vehicles and the target, which can be expressed as:
\begin{equation}
\mathcal{S}=\{\mathcal{S}^{vehicle},\mathcal{S}^{target}\}.
\end{equation} 

The state of vehicles \(\mathcal{S}^{vehicle}\) includes the kinematic features of both the ego vehicle (EV) and surrounding vehicles (SVs), which is represented by an array:
\begin{equation}
\mathcal{S}^{vehicle} = \{s_i = (x, y, v^x, v^y, \varphi)_i \mid i \in \{0, 1, 2, \ldots, N\}\},
\end{equation} 
where \(s_0\) represents the information of the EV, and \(s_i\) for \(1 \leq i \leq N\) denotes the information of the \(N\) SVs. \(x, y\) and \(v^x,v^y\) represent the position and speed in longitudinal and lateral directions, respectively. \(\varphi\) represents the heading angle. The features of SVs are calculated using a coordinate system relative to the EV, while EV remains in an global coordinate system. To ensure the stability and accelerate the learning process, all observations are normalized within a fixed range. 

The state of target \(\mathcal{S}^{target}\) includes the position information of the target. It is formulated as:
\begin{subequations}\label{eq:sgoal}
\begin{align}
& \mathcal{S}^{target} = \{d, l\}, \\
& d = \|\mathbf{P}_{ego} - \mathbf{P}_{target}\|_2,\\
& l = \begin{cases}
[1,0,0] & \text{if target on the left lane,}\\
[0,1,0] & \text{if target on the same lane,}\\
[0,0,1] & \text{if target on the right lane,}\\
\end{cases}
\end{align}
\end{subequations}
where \(d\) represents the Euclidean distance between the EV and the target. The three-dimensional one-hot vector \(l\) encodes the relative lane position of the target with respect to the EV's current lane.

2) Action space. The action space comprises five discrete actions that capture fundamental driving maneuvers and speed adjustments, including left lane change, idle, right lane change, accelerate, and decelerate. The detailed description of different actions within the action space is illustrated in Table \ref{tab:actionspace}.
\begin{table}[h]
\caption{Defined Action Space\label{tab:actionspace}}
\centering
\begin{tabular}{ccc}
\hline
Action                 & Driving Behavior                                       \\ \hline
Left lane change       & Left lane change with speed unchanged                    \\
Idle                   & Keep lane with speed unchanged                         \\
Right lane change      & Right lane change with speed unchanged                   \\ 
Accelerate             & Increase target speed by 1 m/s                         \\ 
Decelerate             & Decrease target speed by 1 m/s                       \\ \hline
\end{tabular}
\end{table}

3) Reward function. The proposed reward function consists of four components: mission success, lane change efficiency, driving speed, and driving safety. The total reward $R$ is formulated as:
\begin{equation}\label{eq:total_reward}
    R = R_{mission} + R_{lanechange} + R_{speed} + R_{safety}
\end{equation}

$R_{mission}$ is designed to incentivize mission success while penalizing mission failure, as defined by:
\begin{equation}\label{eq:mission_reward}
R_{mission} = \begin{cases} 
    \omega_1 & \text{if mission success,}\\
    -\omega_2 & \text{if mission failure,}\\
    0 & \text{otherwise,} 
\end{cases}
\end{equation}
where $\omega_1$ and $\omega_2$ are positive weight coefficients.

$R_{lanechange}$ encourages appropriate lane change behaviors, defined as follows:
\begin{equation}\label{eq:lanechange_reward}
R_{lanechange} = \begin{cases} 
    \omega_3 & \text{if } \varphi_{0} > \epsilon_1, \\
    -\omega_4 & \text{if } \varphi_{0} < -\epsilon_1, \\
    0 & \text{otherwise,}
\end{cases}
\end{equation}
where $\omega_3$ and $\omega_4$ are positive weight parameters designed to promote rightward lane change and penalizes leftward lane change. $\varphi_{0}$ represents the heading angle of EV, and $\epsilon_1$ serves as a small threshold to mitigate signal noise.

$R_{speed}$ encourages maintaining the highest speed within the lane's speed limit. It is formulated as:
\begin{equation}\label{eq:speed_reward}
R_{speed} = \begin{cases} 
\frac{\omega_5 \cdot (v_{ego} - v_{min})}{v_{max} - v_{min}} & \text{if } v_{ego} \in [v_{min}, v_{max}], \\
0 & \text{otherwise,} 
\end{cases}
\end{equation}
where \(v_{ego}\) represents the speed of EV. \(v_{max}\) and \(v_{min}\) denote the maximum and minimum speed of the lane. \(\omega_5\) is a weight parameter larger than 0.

$R_{safety}$ is a penalty term based on the Time-to-Collision (TTC) with the nearest front vehicle, defined as follows:
\begin{subequations}\label{eq:safety_reward}
\begin{align}
& R_{safety} = \begin{cases}
    \frac{-\omega_6}{\max(0.1, \text{TTC}(ego, front))} & \text{if } \frac{\|\mathbf{P}_{ego} - \mathbf{P}_{front}\|_2}{v_{ego}} < 1, \\
    0 & \text{otherwise,}
\end{cases} \\
& \text{TTC}(ego, front) = \begin{cases}
\frac{\|\mathbf{P}_{ego} - \mathbf{P}_{front}\|_2}{v_{ego} - v_{front}} & \text{if } v_{ego} > v_{front}, \\
\infty & \text{otherwise,}
\end{cases}
\end{align}
\end{subequations}
where $\|\mathbf{P}_{ego} - \mathbf{P}_{front}\|_2$ represents the Euclidean distance between the ego vehicle and the nearest front vehicle, $v_{front}$ is the nearest front vehicle's speed, and $\omega_6$ is a positive weight parameter that penalizes potential collision risks.

\begin{figure}[t]
\centering
\includegraphics[width=0.99\linewidth]{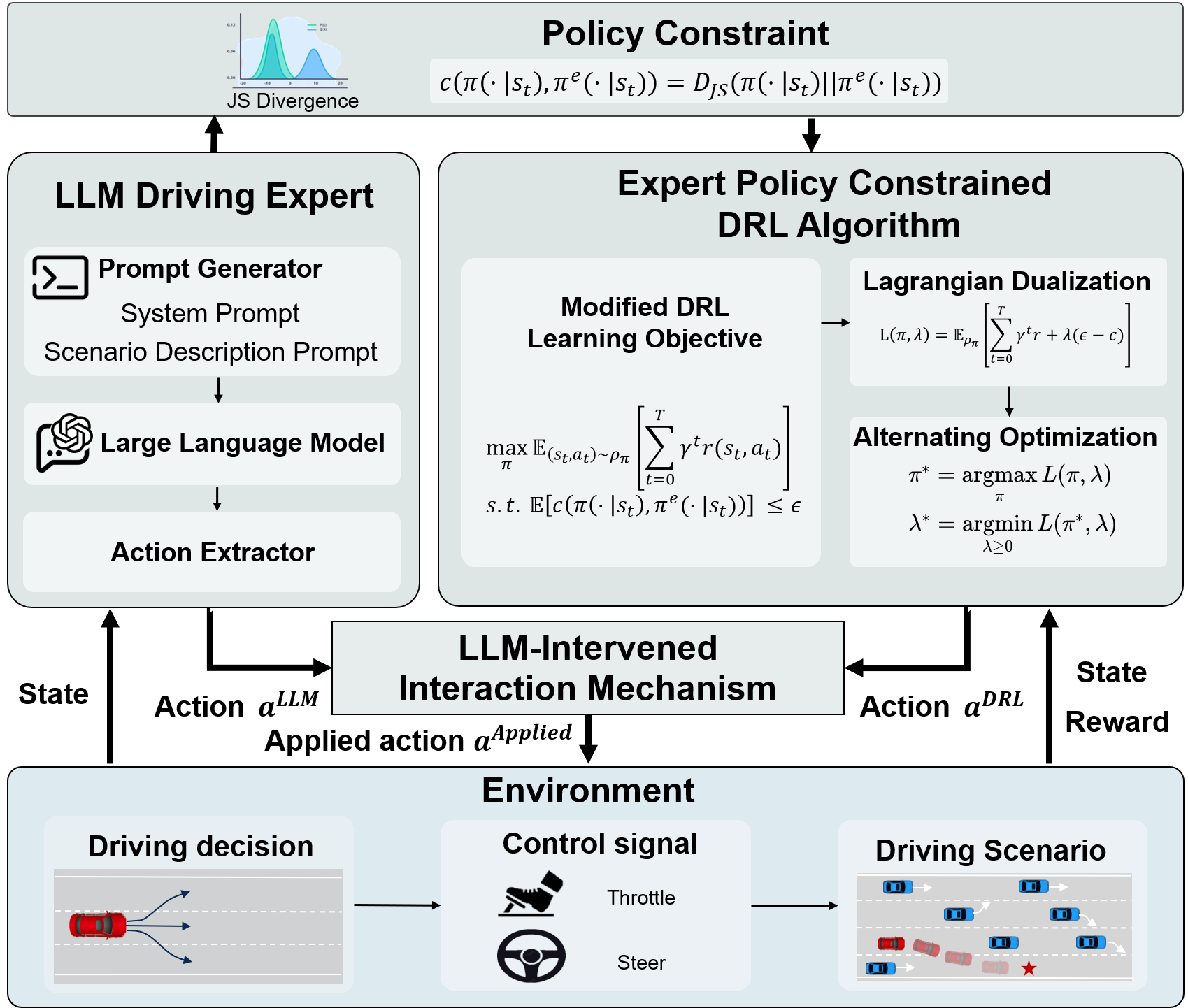}
\caption{LLM guided deep reinforcement learning framework. Within this framework, an LLM driving expert is prompted to guide the learning process of the DRL agent. A novel expert policy constrained DRL algorithm, which integrates a policy constraint based on Jensen-Shannon (JS) divergence into the learning objective, is used to facilitate the DRL agent to learn more effectively from the expert guidance. The actions applied to the environment are determined by a novel LLM-intervened interaction mechanism, which allows the LLM expert to intervene in the DRL agent actions when necessary.}
\label{fig:framework}
\end{figure}

\section{LLM Guided Deep Reinforcement Learning}
The proposed framework is depicted in Fig. \ref{fig:framework}. First, we construct an LLM driving expert to provide action guidance during the DRL agent's learning process. Then, an expert policy constrained DRL algorithm is proposed. This algorithm incorporates the JS divergence policy constraint into its learning objective to regularize the DRL policy to more similar to the LLM expert policy, thereby facilitating the efficient integration of the LLM expert's prior knowledge into the DRL agent's learning process. Furthermore, a novel LLM-intervened interaction mechanism is used to substitute the DRL agent’s catastrophic actions with the guidance action provided by the LLM expert to interact with the environment. Finally, the environment translates the discrete driving decisions into specific control signals, enabling state transitions within the environment and providing corresponding rewards. 

\begin{figure}[t]
\centering
\includegraphics[width=0.99\linewidth]{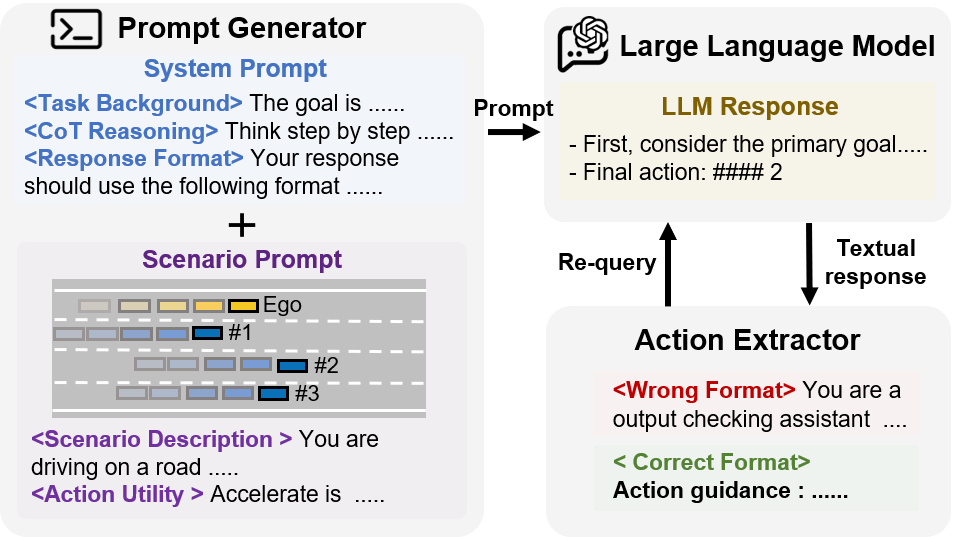}
\caption{The LLM generates a textual response based on the prompts created by the prompt generator. The action extractor then extracts the corresponding action guidance from this response. A re-query mechanism within the action extractor is used to revise the response into a correct format.}
\label{fig:llm_workflow}
\end{figure}

\subsection{LLM Driving Expert}
As shown in Fig. \ref{fig:llm_workflow}, the LLM driving expert comprises three essential components: a prompt generator, an out-of-the-box LLM, and an action extractor. Initially, the prompt generator formulates a prompt based on the current driving scenario. Subsequently, the LLM driving expert processes this prompt to analyze the driving scenario and produces a corresponding response. Finally, the action extractor interprets and decodes the action guidance embedded within the response. 

The prompt generator consists of the system prompt and the scenario prompt. The system prompt sets the context for the LLM by providing necessary background information. In the system prompt, we employ the CoT prompting technique, which guides the LLM to decompose problems into a series of logical steps \cite{cot}. This method mirrors human problem-solving strategies, where intermediate steps and sub-decisions are articulated to reach a final conclusion. 

The scenario prompt offers a comprehensive overview of the current driving environment. An environment context encoder uses structured language to translate the observations from the environment into textual scenario descriptions. This encoder captures the kinematic states of the EV and SVs, as well as static lane information and target information.

The action extractor converts the textual response generated by the LLM into action guidance. Firstly, it parses the response and extracts specific action guidance. In cases where the response is incorrectly formatted, the action extractor initiates a re-query mechanism. This process prompts the LLM to provide a revised response that adheres to the required format.

\subsection{DRL with Policy Constraint} 
To effectively utilize the expert guidance in policy optimization, the policy constraint explicitly limits the deviation between the DRL policy and the expert policy. Specifically, the policy constraint serves as a mechanism that restricts the DRL agent's policy search space to a high-quality region defined by the expert policy, thereby enhancing the learning efficiency and the performance of the derived DRL policy.

The policy constraint is then incorporated into the learning objective of DRL. Hence, the learning objective of DRL with policy constraint is to maximize the cumulative reward while adhering to the expert policy constraint, which can be formulated as a constrained optimization problem:
\begin{equation}\label{eq:obj_1}
\begin{aligned}
&\mathop{\max}_{\pi} J(\pi) =  \mathop{\max}_{\pi} \mathbb{E}_{(s_t,a_t) \sim \rho_{\pi}}\left[\sum_{t=0}^{T} \gamma^t r(s_t, a_t)\right], \\
&\text{s.t.} \quad \mathbb{E}\left[c\left(\pi(\cdot|s_t),\pi^e(\cdot|s_t)\right)\right] \le \epsilon,
\end{aligned}
\end{equation}
where \(c(\cdot)\) denotes the constraint function. \(\pi(\cdot|s_t)\) and \(\pi^{e}(\cdot|s_t)\) represent the DRL policy and the expert policy. \(\epsilon\) represents the maximum policy deviation, which is smaller than 1. 

The constraint function is mathematically represented by the JS divergence between the DRL policy and the LLM expert policy. The JS divergence is a symmetrized and normalized variant of the Kullback-Leibler (KL) divergence, offering a bounded measure of similarity between two probability distributions, ranging from 0 to 1. By employing the JS divergence, we quantify the divergence between the DRL policy and the expert policy. The constraint function can be expressed as:
\begin{equation}\label{eq:constrain_1}
c\left(\pi(\cdot|s_t),\pi^{e}(\cdot|s_t)\right) = D_{JS}(\pi(\cdot|s_t) \parallel \pi^{e}(\cdot|s_t)),
\end{equation}
where \(D_{JS}\) signifies the JS divergence, which is defined as:
\begin{equation}\label{eq:constrain_2}
D_{JS}(\pi \parallel \pi^e) = \frac{1}{2} \left[ D_{KL}(\pi \parallel \overline{\pi}) + D_{KL}(\pi^e \parallel \overline{\pi}) \right],
\end{equation}
where \(D_{KL}\) represents the KL divergence. \(\overline{\pi}\) is the average policy of the DRL and the expert policy. It is defined as:
\begin{equation}\label{eq:constrain_3}
\overline{\pi} = \frac{1}{2} \left( \pi + \pi^e \right).
\end{equation}

Therefore, the expert policy constraint is formalized as:
\begin{equation}
\begin{aligned}\label{eq:constrain_5}
c\left(\pi(\cdot|s_t),\pi^{e}(\cdot|s_t)\right) = &\frac{1}{2} \sum_{a \in \mathcal{A}} \pi(\cdot|s_t) \log \frac{\pi(\cdot|s_t)}{\overline{\pi}(\cdot|s_t)}\\
+ &\frac{1}{2}\sum_{a \in \mathcal{A}} \pi^e(\cdot|s_t) \log \frac{\pi^e(\cdot|s_t)}{\overline{\pi}(\cdot|s_t)}.
\end{aligned}
\end{equation}

\subsection{Expert Policy Constrained DRL Algorithm} 

The expert policy constrained DRL algorithm incorporates the policy constraint into the actor-critic framework to solve the constrained optimization problem (\ref{eq:obj_1}). Based on Lagrangian duality theory \cite{lag}, this problem is first transformed into its Lagrangian dualization form, which is formulated as follows:
\begin{equation}\label{eq:lag}
\begin{aligned}
L(\pi,\lambda) = \mathbb{E}_{(s_t,a_t) \sim \rho_{\pi}}\left [\sum_{t=0}^{T}\gamma^t r(s_t,a_t) + \lambda(\epsilon-c(\pi,\pi^e)\right ],
\end{aligned}
\end{equation}
where \(\lambda\) is a dual variable. Assuming strong duality, the original problem can be transformed into the following Lagrangian dual problem:
\begin{equation}\label{eq:obj_lag}
\mathop{\max}_{\pi} J(\pi) = \mathop{\min}_{\lambda \ge 0}\mathop{\max}_{\pi}L(\pi,\lambda).
\end{equation}

The optimal policy \(\pi^*\) and the optimal dual variable \(\lambda^*\) can be determined through an alternating optimization process. First, by fixing the dual variable \(\lambda\), we maximize the Lagrange function \(L(\pi,\lambda)\) to find the optimal policy \(\pi^*\). Subsequently, we substitute the optimal policy back in and minimize \(L(\pi^*,\lambda)\) to compute the optimal dual variable \(\lambda^*\). This process is formally described as follows:
\begin{subequations}
\begin{align}
&\pi^* = \mathop{argmax}_{\pi}L(\pi,\lambda)\label{eq:pi_lag}, \\
&\lambda^* = \mathop{argmin}_{\lambda \ge 0}L(\pi^*,\lambda). \label{eq:lambda_lag}
\end{align}
\end{subequations}

To solve for the optimal policy \(\pi^*\), we employ a policy iteration (PI) scheme. PI primarily consists of two processes: policy evaluation and policy improvement.

1) The policy evaluation process is carried out through the critic network. The critic network is used to estimate the action value function \(Q(s, a)\), which represents the expected state value taking action \(a\) in state \(s\). The optimal action value function \(Q^*(s,a)\) can be learned through repeatedly applying a modified Bellman backup operator \(\mathcal{T}\) as formalized below:
\begin{equation}\label{eq:Q}
\mathcal{T}Q(s_t,a_t):=r(s_t,a_t)+\gamma\mathbb{E}_{s_{t+1}\sim p(s_t)}\left[ V(s_{t+1})\right],
\end{equation}
where \(V(s_{t+1})\) represents the state value function under state \(s_{t+1}\), which is defined as follows:
\begin{equation}\label{eq:V}
    V(s_{t+1})= \mathbb{E}_{a_{t+1}\sim\pi_{\theta}(s_{t+1})}[Q(s_{t+1},a_{t+1})-\lambda c(\pi_{\theta},\pi^e)],
\end{equation}
Here, \(\pi_{\theta}\) represents the DRL policy, which is parameterized by the actor network. \(\theta\) represents the actor network's parameters. To mitigate the impact of overestimation and enhance the stability of the algorithm, the critic network comprises two independent neural networks, denoted as \(Q_{\phi_z},z \in \{1, 2\}\), where \(\phi_z\) represents the critic network's parameters. Additionally, two target critic networks \(\bar{Q}_{\bar{\phi}_{z}}(s,a)\) sharing the same parameters as \(Q_{\phi_{z}}(s,a)\) are utilized. The parameters of \(\bar{Q}_{\bar{\phi}_{z}}(s,a)\) is updated via Polyak averaging as follows:
\begin{equation}\label{eq:targetupdate}
    \bar{\phi}_z \leftarrow  \tau\bar{\phi}_z + (1-\tau)\phi_z,
\end{equation}
where \(\tau\) is a soft update factor between 0 and 1.

The loss function for the critic network is defined by the Bellman residual, which is represented as follows:
\begin{equation}\label{eq:c_loss}
    J_c(\phi_z)= \frac{1}{N} \mathop{\mathbb{E}}\limits_{Ts\sim \mathcal{D}} \left[ \left(\mathcal{T}\bar{Q}_{\bar{\phi}_z}(s,a) - Q_{\phi_{z}}(s,a)\right)^2 \right],
\end{equation}
where \(Ts = (s,a,r,s') \) represents the transition tuples and \(\mathcal{D}\) denotes the replay buffer, with \(N\) being the number of samples. 

According to (\ref{eq:Q}) and (\ref{eq:V}), \(\mathcal{T}\bar{Q}_{\bar{\phi}_z}\) is defined as:
\begin{equation}
\begin{cases}
\begin{aligned}
&\mathcal{T}\bar{Q}_{\bar{\phi}_z}(s, a) = r(s, a) + \gamma\mathbb{E}_{s'\sim p(s)}[V(s')],\\
&V(s') = \mathbb{E}_{a' \sim \pi_{\theta}(s')} \left[ \min_{z \in \{1, 2\}} \bar{Q}_{\bar{\phi}_z}(s', a') - \lambda c(\pi_{\theta}, \pi^e) \right].
\end{aligned}
\end{cases}
\end{equation}

2) The policy improvement process is used to optimize the DRL policy \(\pi_{\theta}\), which is carried out through updating the actor network. The actor network outputs the probability of each action being selected by the DRL agent based on the input state vector. From (\ref{eq:pi_lag}), the optimal policy \(\pi_{\theta}^*\) is determined by maximizing \(L(\pi_{\theta},\lambda)\), which can be expressed as follows:
\begin{equation}\label{eq:pi_loss}
\begin{aligned}
\pi_{\theta}^* &= \mathop{\arg\max}_{\pi_{\theta}} L(\pi_{\theta},\lambda) \\
      &= \mathop{\arg\max}_{\pi_{\theta}} \mathbb{E}_{(s_t,a_t) \sim \rho_{\pi_{\theta}}}\left [\sum_{t=0}^{T}\gamma^t r(s_t,a_t) - \lambda c(\pi_{\theta},\pi^e)\right ].
\end{aligned}
\end{equation}

In (\ref{eq:pi_loss}), the \(\lambda \cdot \epsilon\) term is omitted because, for a fixed \(\lambda\), it is a constant and does not affect the gradient of the actor network. The cumulative reward is estimated through the critic network above. Therefore, the parameters of the actor network can be updated by maximizing the following objective function:
\begin{equation}\label{eq:a_loss}
J_a(\theta) = \frac{1}{N}\mathop{\mathbb{E}}\limits_{Ts\sim D}
\left[\sum\limits_{\substack{a\in\mathcal{A}\\a\sim\pi_{\theta}(s)}}\left(\mathop{min}\limits_{z\in{1,2}}Q_{\phi_z}(\cdot,s)-\lambda c(\pi_{\theta},\pi^e) \right)\right].
\end{equation}

The dual variable \(\lambda\) is optimized by minimizing the Lagrange function \(L(\pi_{\theta}^*, \lambda)\). This can be achieved by minimizing the following loss function:
\begin{equation}\label{eq:lambda_loss}
J(\lambda)=\frac{1}{N}\mathop{\mathbb{E}}\limits_{Ts\sim D}\left[\sum\limits_{\substack{a\in\mathcal{A}\\a\sim\pi_{\theta}(s)}}\lambda(\epsilon-c(\pi^*_{\theta},\pi^e) \right].
\end{equation}

The dual variable \(\lambda\) is updated by:
\begin{equation}\label{eq:lambda_update}
\lambda \leftarrow \lambda - \alpha(\epsilon-c(\pi^*_{\theta},\pi^e)),
\end{equation}
where \(\lambda\) is the learning rate. According to (\ref{eq:lambda_update}), when the policy divergence \(c(\pi^*_{\theta},\pi^e)\) between DRL policy and expert policy exceeds the policy deviation threshold \(\epsilon\), the dual variable \(\lambda\) increases, making the DRL agent places greater emphasis on aligning its policy with the expert policy. Conversely, \(\lambda\) decreases when the divergence falls below \(\epsilon\), allowing the agent to prioritize reward optimization. Algorithm \ref{alg:EGC-AC} outlines the expert policy constrained actor-critic algorithm in detail.

\begin{algorithm}[h]
\caption{Expert Policy Constrained DRL Algorithm}\label{alg:EGC-AC}
\begin{algorithmic}
\STATE \textbf{Input:} maximum episode number $M$, batch size $N$, empty replay buffer $\mathcal{D}$, expected minimum policy deviation $\epsilon$.
\STATE Initialize actor network parameters $\theta$.
\STATE Initialize critic network parameters $\phi_z$. 
\STATE Initialize dual variable $\lambda$.
\STATE Initialize target network parameters $\bar{\phi}_1 \leftarrow \phi_1, \bar{\phi}_2 \leftarrow \phi_2$.
\FOR{episode = 1 to $M$}
    \STATE Reset environment state $s_0$.
    \WHILE{environment state is not terminal}
        \STATE DRL agent select action $a_t^{\text{DRL}}\sim\pi_\theta$.
        \STATE LLM expert provides guidance action $a_t^{\text{LLM}}\sim\pi^e$.
        \STATE Obtain applied action $a_t^{Applied}$ through Eq. (\ref{eq:safetychecker}).
        \STATE Apply action $a_t^{Applied}$ to environment. Obtain reward $r_t$ and next state $s_{t+1}$.
        \STATE Store the transition in the replay buffer $\mathcal{D}$.
        \STATE Sample $N$ transitions from $\mathcal{D}$.
        \STATE Update the critic network parameters through Eq. (\ref{eq:c_loss}):\\$\phi_z \leftarrow \phi_z - \alpha \nabla J_c(\phi_z)$.
        \STATE Update the actor network parameters through Eq. (\ref{eq:a_loss}):\\$\theta \leftarrow \theta - \alpha \nabla J_a(\theta)$.
        \STATE Update the dual variables through Eq. (\ref{eq:lambda_loss}):\\$\lambda \leftarrow \lambda - \alpha \nabla J(\lambda)$.
        \STATE Update the target network parameters through Eq. (\ref{eq:targetupdate}):\\$\bar{\phi}_z \leftarrow  \tau\bar{\phi}_z + (1-\tau)\phi_z$.    
    \ENDWHILE
\ENDFOR
\end{algorithmic}
\end{algorithm}

\subsection{Expert-Intervened Interaction Mechanism}
In the training stage, when the DRL agent interacts with the environment in a standard approach, the actions applied in the environment are DRL actions, which are sampled from the action distribution of DRL policy \(\pi_{\theta}\):
\begin{equation}\label{eq:arl}
a^{DRL}_t \sim \pi_{\theta}(\cdot|s_t),
\end{equation}
however, this standard interaction approach often results in catastrophic actions, impeding the agent to acquire sufficient successful trajectories and hindering policy optimization. 

\begin{figure}[t]
\centering
\includegraphics[width=0.97\linewidth]{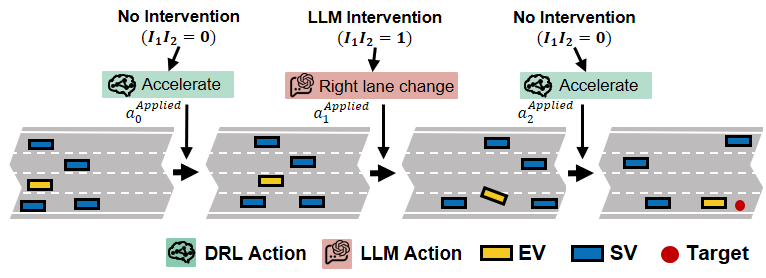}
\caption{The expert-intervened interaction mechanism allows the LLM expert to intermittently intervene in the interactions between the DRL agent and the environment based on the DRL action safety condition and the intervention permission condition.}
\label{fig:intervenpic}
\end{figure}

In order to effectively utilize expert guidance to mitigate these limitations, a novel expert-intervened interaction mechanism is proposed to replace the standard interaction approach. As shown in Fig. \ref{fig:intervenpic}, this mechanism allows for intervention by the LLM expert in the interactions between the DRL agent and the environment during training. In this context, intervention refers to substituting the DRL agent's action with the guidance action provided by the LLM expert to interact with the environment. Note that the expert-intervened interaction mechanism is only activated in the training stage. In the testing stage, the LLM expert does not intervene in the interaction between the DRL agent and the environment. Specifically, with the expert intervened interaction mechanism, the action applied to the environment are determined based on the DRL action safety condition and intervention permission condition. The final applied action \(a_t^{Applied}\) can be expressed as follows:
\begin{equation}\label{eq:safetychecker}
a^{Applied}_t = \left(1-I_1I_2\right)\cdot a_t^{DRL}+I_1I_2\cdot a_t^{LLM},
\end{equation}
where \(a_t^{DRL}\) and \(a_t^{LLM}\) represent the action of the DRL agent and the guidance action of the LLM expert, respectively. \(I_1\) is a DRL action safety indicator, and \(I_2\) is an intervention permission indicator. 

The DRL action safety indicator \(I_1\) is set to 1 when the DRL agent outputs a hazardous action, and 0 otherwise. The safety assessment of DRL lane-changing actions considers the TTC between the EV and the front vehicle, rear vehicle in the target lane, denoted as \(T_{ft}\) and \(T_{rt}\), respectively. These TTC values are evaluated against their respective thresholds, \(\tau_{ft}\) and \(\tau_{rt}\). Similarly, for the safety verification of other actions, the TTC values \(T_{f}\) and \(T_{r}\) between the EV and the front vehicle, rear vehicle in the current lane are compared against their respective thresholds \(\tau_{f}\) and \(\tau_{r}\). 

The intervention permission indicator \(I_2\) is used to judge whether the LLM expert is allowed to intervene in the interaction between the DRL and the environment during the current training episode, with a value of 1 indicating permission and 0 indicating otherwise. In our proposed expert-intervened interaction mechanism, the LLM expert intervenes in an intermittent mode. Specifically, several training episodes are randomly chosen as the intervention permitted episodes set \(\mathcal{C}\). In the remaining episodes, the LLM expert is not allowed to intervene.
The expert-intervened interaction mechanism is illustrated in Algorithm \ref{alg:safetychecker}.

\begin{algorithm}[t]
\caption{Expert-intervened Interaction Mechanism}\label{alg:safetychecker}
\begin{algorithmic}
\STATE \textbf{Input:} Driving state $s_t$, DRL agent action $a_t^{\text{DRL}}$, LLM expert guidance action $a_t^{\text{LLM}}$, Intervention permitted episodes set $\mathcal{C}$, Episode $E$.
\IF{$a_t^{\text{DRL}} \in \{\text{Left lane change, Right lane change}\}$}
    \STATE Compute $T_{ft}, T_{rt}$ based on driving state $s_t$.
    \STATE $I_1 \gets \left[(T_{ft} < \tau_{ft}) \lor (T_{rt} < \tau_{rt})\right]$.
\ELSIF{$a_t^{\text{DRL}} \in \{\text{Accelerate, Idle, Decelerate}\}$}
    \STATE Compute $T_{f}, T_{r}$ based on driving state $s_t$.
    \STATE $I_1 \gets \left[(T_{f} < \tau_{f}) \lor (T_{r} < \tau_{r})\right]$.
\ENDIF
\IF{$E\in\mathcal{C}$}
    \STATE Set $\mathit{I_2}=1$.
\ELSE
    \STATE Set $\mathit{I_2}=0$.
\ENDIF
\STATE Compute the applied action $a^{Applied}_t$ through Eq. (\ref{eq:safetychecker}).
\STATE \textbf{Output:} Applied action $a^{Applied}_t$.
\end{algorithmic}
\label{alg1}
\end{algorithm}

\section{Experimental Setup}
We first introduce the experimental scenario. Then several baseline methods are implemented for comparison. Finally, we elaborate the implementation details of the DRL algorithms.
\begin{figure}[h]
\centering
\includegraphics[width=0.97\linewidth]{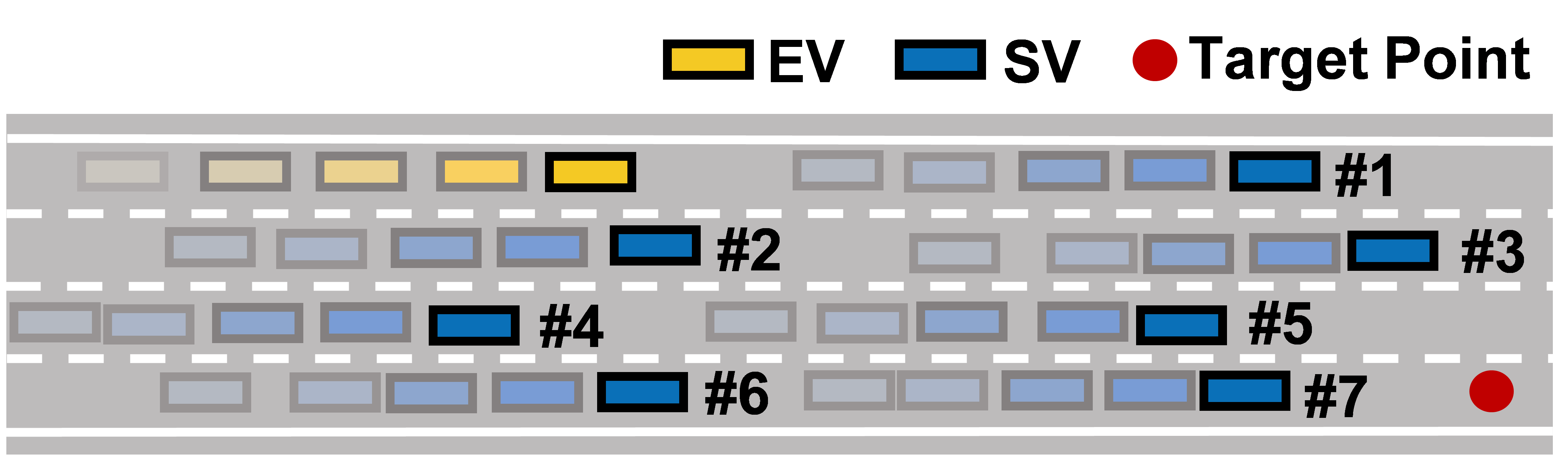}
\caption{The experimental scenario constructed by the highway-env simulator. The yellow vehicle represents the ego vehicle, the blue vehicles represent the surrounding vehicles. The shading represents the historical trajectory of the vehicles. The red dot represents the target point.}
\label{fig:scenario}
\end{figure}

\subsection{Experimental Scenarios}
We utilize the highway-env simulator to construct the experimental scenario \cite{highwayenv}. As illustrated in Fig. \ref{fig:scenario}, the experimental scenario consists of four driving lanes, each measuring 1000 m in length and 4 m in width, with a speed limit of 30 m/s. All vehicles in the simulation have a length of 5 m and a width of 2 m. The simulation runs for 20 s per episode and one time step is 0.05 s. At the beginning of each episode, the EV is initialized at a random position in the leftmost lane with an initial speed of 20 m/s. The target point is positioned 500 m away in the rightmost lane. The EV must implement actions to successfully reach the target point within a limited time period and avoid collisions. The SVs are generated at random positions around the EV. The number of SVs is 30. The initial distances between these SVs range from 20 m to 70 m. The initial speeds of the SVs are between 20 m/s and 25 m/s, with a maximum speed of 30 m/s. The main parameters of the experimental scenario is listed in Table \ref{tab:simulation}.

\begin{table}[t]
\centering
\caption{Main Parameters of the Experimental Scenario\label{tab:simulation}}
\begin{tabular}{ccc}
\hline
Parameter                    & Details                     \\ \hline
EV Initial Speed             & 20 m/s                               \\ 
EV Initial Position          & Leftmost lane                        \\
SV Number                    & 30                                   \\ 
SV Initial Speed             & 20 m/s to 25 m/s                     \\
SV Initial Position          & Random positions around EV          \\
Vehicle Size                 & Length: 5 m, Width: 2 m              \\ 
Driving Lanes                & Length: 1000 m; Width: 4 m          \\ 
Target Point                 & Rightmost lane, 500 m from EV        \\ 
Time Step                    & 0.05 s                               \\ 
Simulation Duration          & 20 s per episode                     \\ \hline
\end{tabular}
\end{table}

\begin{table}[t]
\caption{Illustration of the Baseline Methods \label{tab:baseline}}
\centering
\begin{tabular}{cccc}
\hline
Methods           &\makecell{Expert\\guidance}    &\makecell{Expert guidance\\utilization method}   &\makecell{Expert\\intervention}   \\ \hline
Vanilla-SAC       &N/A                 &N/A                                      & N/A \\
SAC+RP            &Online             &Reward penalty                           & Yes \\
SAC+BC            &Online             &Behavior clone loss                      & Yes \\
SAC+Demo          &Offline            &Margin classification loss               & No \\
\hline
\end{tabular}
\end{table}

\subsection{Comparison Baselines} 
We implement several state-of-the-art DRL methods as baselines to  comparatively evaluate our proposed method. The baseline algorithm are listed as follows.

1) Vanilla-SAC \cite{sac}: As a state-of-the-art off-policy DRL algorithm, SAC achieves policy optimization by maximizing the expected cumulative reward and entropy simultaneously. The vanilla-SAC algorithm without any expert guidance is implemented as the baseline for standard DRL.

2) SAC+RP \cite{ir10}: SAC+RP adds an additional reward penalty (RP) term to the original rewards when expert intervention occurs. The RP aims to minimize the frequency of expert interventions. SAC+RP incorporates this penalty mechanism into the SAC algorithm.

3) SAC+BC \cite{bc}: This approach utilizes expert guidance by incorporating a normalized behavior cloning (BC) objective into the original actor network loss function. We modify the actor network of SAC algorithm to implement SAC+BC.

4) SAC+Demo \cite{sacdqfd}. SAC+Demo utilizes expert demonstration (Demo) data which is collected before training to enhance the learning process of DRL. Demonstration data generated by human in the simulator is pre-stored in the replay buffer. In the DRL training stage, both human demonstration data and DRL self-exploration data are used. Moreover, we add a margin classification loss to the critic network loss function to implement SAC+Demo.

These baseline methods can be divided into three categories. The first category is Vanilla-SAC, which is trained without any expert guidance. The second category includes SAC+RP, SAC+BC. Both methods are trained with online expert, meaning the expert provides interventions in real-time during the learning process. The third category is SAC+Demo, which is trained with offline expert, which means that it is trained with pre-collected expert demonstration and the expert does not intervene during the learning process. For the fair comparison, the original experts involved in the baseline methods are replaced with the same LLM expert used in LGDRL. Table \ref{tab:baseline} illustrates the baseline methods.

\begin{table}[t]
\caption{Hyperparameters of the DRL Algorithm\label{tab:hyperpara}}
\centering
\begin{tabular}{cccc}
\hline
Parameter          &Description    & Value           \\ \hline
Maximum episode      &Training cutoff episode number & 500 \\
Exploration step   & Number of steps for random exploration & 1e3\\
Replay buffer size   &Capacity of the replay buffer & 4e4 \\
Batch size     &Capacity of minibatch     & 256   \\
Learning rate   & Learning rate of the networks   & 5e-4   \\
Discount factor   & Discount factor of the Bellman equation   & 0.9     \\
Policy deviation & Policy deviation tolerance & 0.1\\
\hline
\end{tabular}
\end{table}

\begin{figure}[t]
\centering
\includegraphics[width=0.99\linewidth]{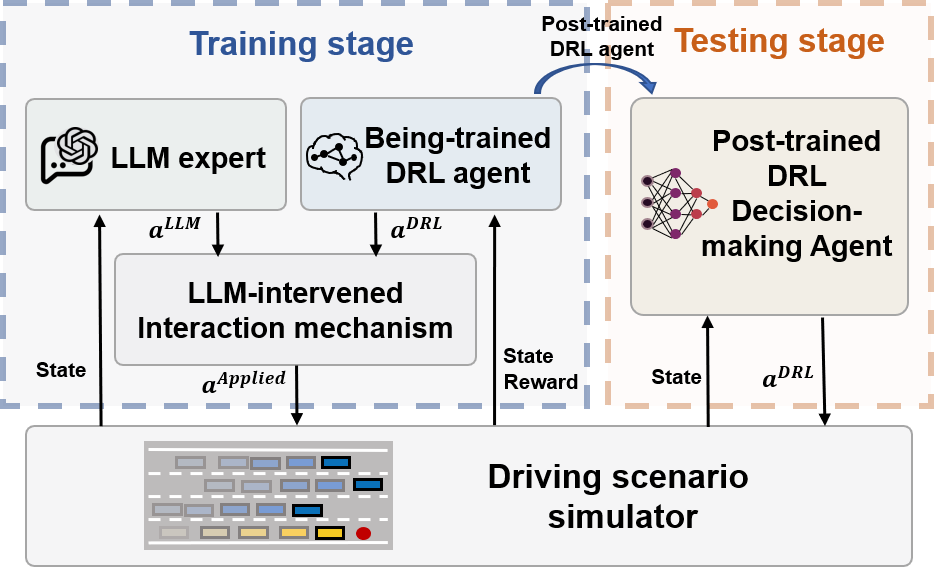}
\caption{Experimental workflows. In the training stage, the DRL agent's learning process is guided by the LLM expert. In the testing stage, the post-trained DRL decision-making agent independently execute the driving tasks without the assistance of the LLM expert.}
\label{fig:traintest}
\end{figure}

\subsection{Implementation Details}
All implemented DRL algorithms have the same neural network architecture. The hidden layers of both the actor and critic networks consist of four fully connected layers (FCN) with 256, 128, 128, and 64 units, respectively. Each FCN employs the ReLU activation function. The primary hyperparameters of the DRL algorithms are detailed in Table \ref{tab:hyperpara}. All DRL algorithms are implemented in PyTorch and optimized with Adam optimizer. The experiment is conducted on a high-performance workstation equipped with an NVIDIA Quadro RTX 8000 GPU. The experimental workflows of the DRL training and testing stage are shown in Fig. \ref{fig:traintest}. In the training stage, the LLM expert, which is constructed based on ChatGPT-4o \cite{llm1}, provides guidance for the DRL agents. In the testing stage, the post-trained DRL agents independently execute the driving tasks without the assistance of the LLM expert.

\section{Result and Discussion}
First, the proposed LGDRL method is compared with the baseline methods in terms of the training and testing performance. Then, we discuss the impact of different intervention modes on the training and testing performance of the proposed method. Finally, we conduct an ablation study to investigate the impact of the expert policy constraint on the proposed method.

\subsection{Training Performance Comparison}

\begin{figure}[t]
\centering
\includegraphics[width=0.99\linewidth]{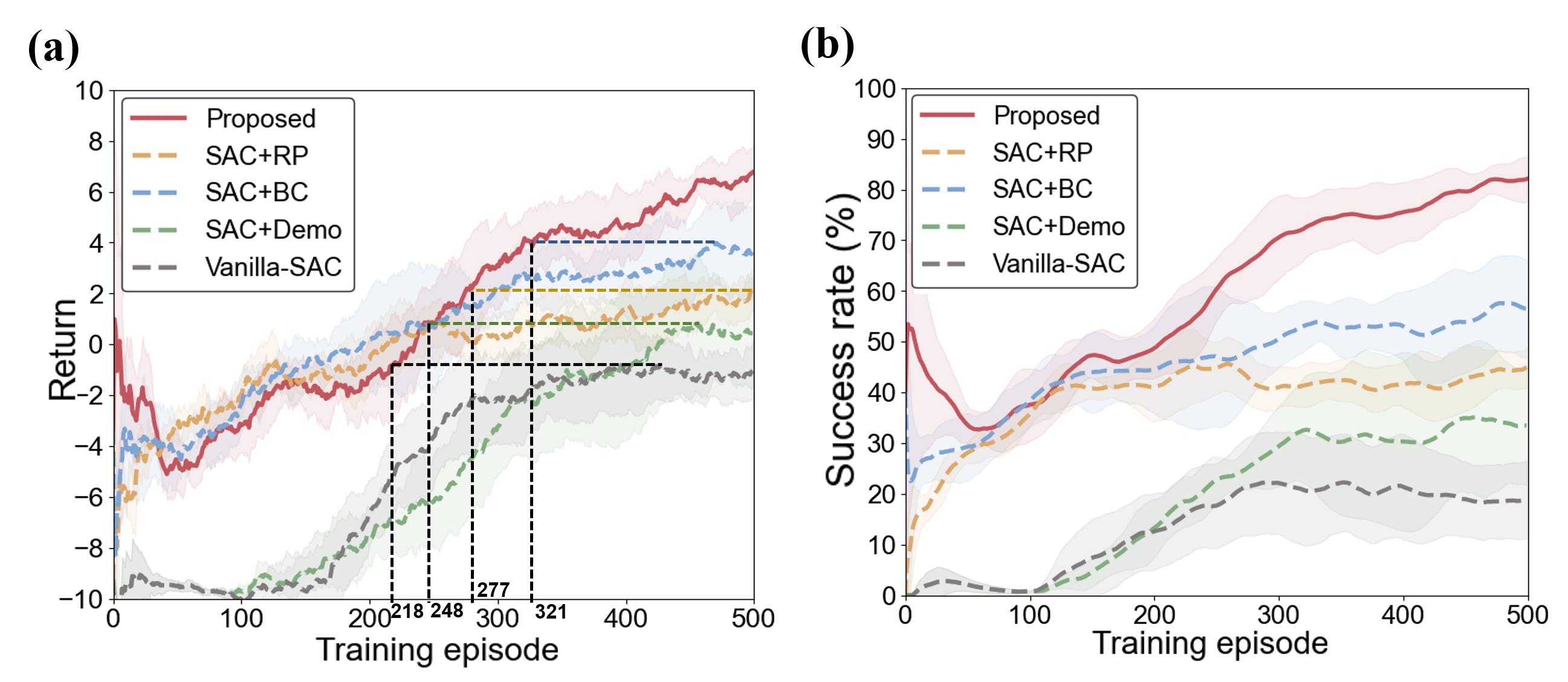}
\caption{Comparison of the training curves of different DRL methods. Based on training data with different random seeds, the mean and standard deviation for each episode are represented by lines and shadows, respectively. (a) The curves of the episode return. (b) The curves of the success rate varying with episodes. The success rate for each episode is calculated based on the results of the most recent 20 episodes.}
\label{fig:train_data}
\end{figure}

The training performance of our proposed LGDRL method is compared with the baseline methods in term of return and success rate. Additionally, we compare the efficiency of expert guidance utilization between LGDRL and baseline methods. To ensure a fair comparison, all DRL methods involving expert guidance use the same LLM expert to provide guidance and all DRL methods are trained five times under the same seed sequence.

Fig. \ref{fig:train_data} illustrates the training performance of different methods through the return and success rate metrics. A higher value of both metrics indicates a better training performance. As can be seen from Fig. \ref{fig:train_data}(a), the Vanilla-SAC method, which is trained without any expert prior knowledge, exhibits the worst asymptotic return. In contrast, the other four methods, which are trained with expert guidance, outperform the Vanilla-SAC method, indicating that DRL can benefit from expert guidance. Among the four expert guidance involved methods, the SAC+Demo method performs worse than the methods trained with online expert guidance. The reason for this result is that online guidance can intervene in the DRL agent's risky actions in real-time, guiding the agent to select actions with higher returns and avoid local optima in policy optimization. Among the three methods with online guidance, the proposed method shows the best performance, followed by SAC+BC and SAC+RP. 

In terms of learning efficiency, the proposed LGDRL method achieves the maximum returns of the Vanilla-SAC and SAC+Demo methods within only 218 and 248 episodes, respectively, demonstrating efficiency improvements of 95.0\% and 83.9\%. Compared to the SAC+RP and SAC+BC methods, LGDRL reached their maximum returns in just 277 and 321 episodes, representing efficiency improvements of 80.5\% and 46.7\%, respectively.

\begin{figure}[t]
\centering
\includegraphics[width=0.99\linewidth]{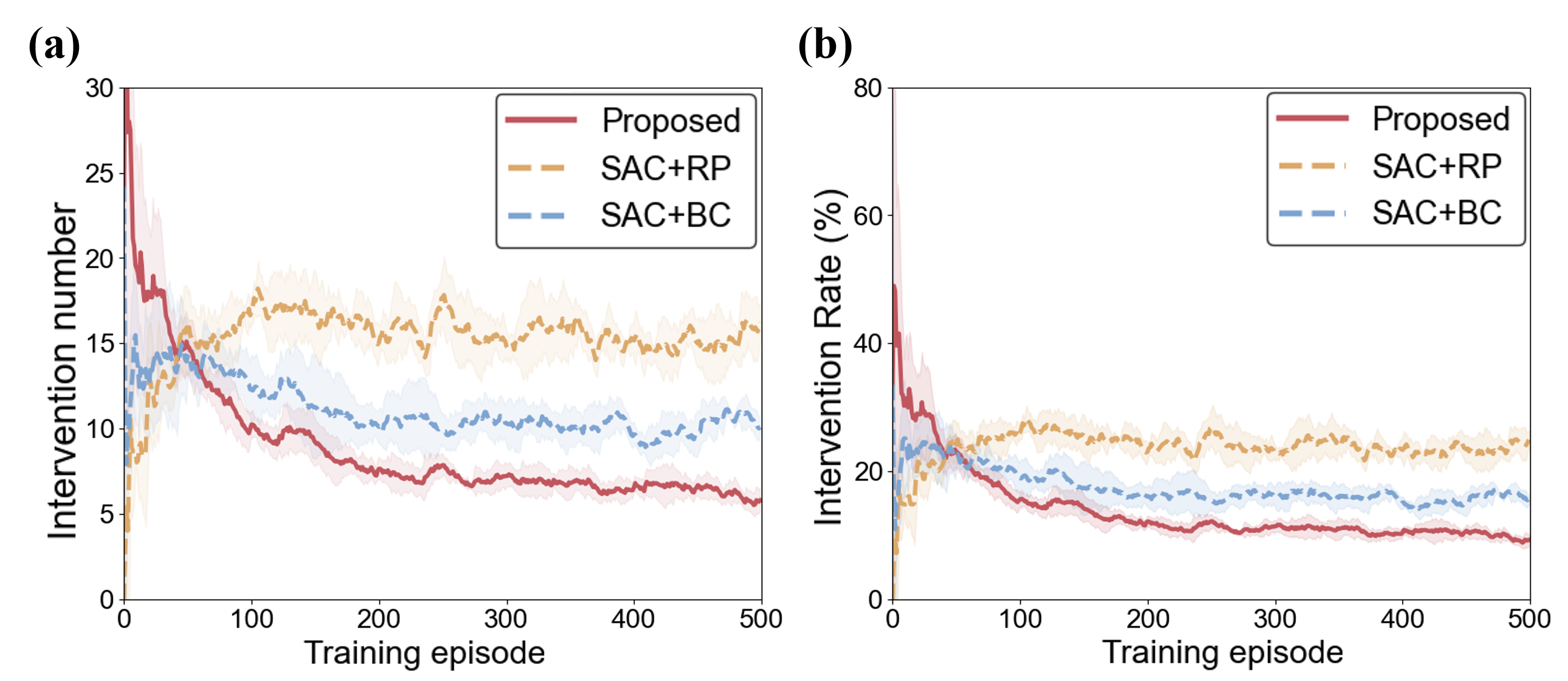}
\caption{Comparison of intervention metrics among different DRL methods with online guidance. The mean values and standard deviation for each episode are represented by lines and shadows. (a) Number of interventions per episode. (b) Intervention rate per episode. The intervention rate is defined as the ratio between the number of time steps with expert intervention and the total number of time steps in each training episode.}
\label{fig:train_interv}
\end{figure}

The asymptotic success rate is selected as the metric to evaluate the final training performance. Fig. \ref{fig:train_data}(b) shows that the proposed LGDRL method achieves the highest performance, with an asymptotic success rate of 82\%, while the Vanilla-SAC and SAC+Demo methods are respectively only 19\% and 33\%, and fluctuate significantly. The SAC+RP and SAC+BC methods achieve success rates 45\% and 56\%, which are also remarkably lagging behind the proposed method. 

The SAC+RP, SAC+BC, and proposed LGDRL methods are trained with online expert guidance. However, these methods utilize expert guidance in different way. We further investigate the efficiency of expert guidance utilization among these three methods. Fig. \ref{fig:train_interv} illustrates  the intervention number and the intervention rate per training episode across these methods. Lower intervention number and intervention rate suggest more efficient utilization of expert guidance, demonstrating the methods decrease the dependency on external expert knowledge. The intervention number of the proposed LGDRL method rapidly decreases as training progresses. The SAC+BC method also shows a reduction, but at a slower rate with greater fluctuations. This can be attributed to the fact that proposed method simultaneously modifies the learning objectives of both the critic and actor networks, whereas the SAC+BC method only modifies the actor network's learning objective by directly combining the BC objective with the original learning objective, which leads to inconsistent update paths between the actor and critic networks, thus causing instability and lower efficiency. The intervention number of the SAC+RP method increases in the early training stages and remains stable at around 15 interventions per episode as training progresses. This is linked to the form of expert guidance information. In the SAC+RP method, the expert guides policy updates through a reward penalty signal. This indirect guidance signal makes it difficult to convey targeted prior information, thus resulting in the lowest efficiency in utilizing expert guidance. 

Table \ref{tab:interv} summarizes the total intervention number and intervention rate in the training stage. The total intervention number represents the cumulative expert interventions throughout the training process, while the total intervention rate is calculated as the ratio of total intervention number to the total number of time steps. The proposed method demonstrates superior performance, exhibiting the fewest interventions and the lowest intervention rate, indicating the highest expert guidance utilization efficiency.
\begin{table}[t]
\caption{Intervention Data in Training Stage of Different Methods\label{tab:interv}}
\centering
\begin{tabular}{cccc}
\hline
          & Total intervention Number     & Total intervention Rate(\%) \\ \hline
Proposed  & \textbf{3.87K±0.17K}          & \textbf{14.83±0.93}    \\
SAC+BC    & 5.35K±0.16K                   & 21.50±1.02             \\
SAC+RP    & 7.69K±0.30K                   & 31.43±2.69             \\ \hline
\end{tabular}
\end{table}

\begin{table}[t]
\caption{Testing Performance of Different Post-trained DRL Agents\label{tab:test_performance_1}}
\centering
\begin{tabular}{ccccc}
\hline
            &Success Rate (\%) &Collision Rate (\%)   & Return \\ \hline
Proposed    & \textbf{90}  & \textbf{10}              & \textbf{6.87±8.27}\\
SAC+BC      & 58           & 38                       & 4.46±10.59 \\
SAC+RP      & 38           & 24                       & 1.85±8.88 \\
SAC+Demo    & 52           & 24                       & 3.51±9.55 \\
Vanilla-SAC & 8            & 26                       & -3.57±5.48 \\  \hline
\end{tabular}
\end{table}

\begin{table}[t]
\caption{CPU Clock Time Consumption of Inference per Time Step\label{tab:decisiontime}}
\centering
\begin{tabular}{ccccc}
\hline
            & Time consumption per time step (s)   &  \\ \hline
Proposed    & 0.01                  &\\
LLM Expert  & 3.72                &   \\     \hline
\end{tabular}
\end{table}

\subsection{Testing Performance Comparison}
The testing performance of the post-trained DRL agents are evaluated in this section. All the post-trained DRL agents obtained from different methods are tested for 50 episodes. It is important to note that during the testing stage, all post-trained DRL agents operated without any assistance from the LLM expert. The testing performance of different post-trained DRL agents is evaluated based on success rate, collision rate, and return. The success rate and collision rate refer to the proportion of successful episodes and episodes with collisions out of all test episodes, respectively. These two metrics are used to evaluate the agent's task success rate and driving safety. Return refers to the cumulative reward obtained by the EV in each episode during the testing process, used to evaluate the overall performance. Table \ref{tab:test_performance_1} shows the testing performance of different post-trained DRL agents. Results in Table \ref{tab:test_performance_1} demonstrates that the proposed method achieves a success rate and collision rate of 90\% and 10\%, outperforming other methods. The proposed method also exhibits the highest return. 

Moreover, the inference time consumption of the post-trained LGDRL agent is compared with that of the LLM expert in the testing stage. Both the post-trained LGDRL agent and the LLM expert are evaluated independently for one episode in an identical, randomly generated driving scenario. Table \ref{tab:decisiontime} presents the average CPU clock time consumption for decision inference per time step during this testing episode. The results reveal that the LLM expert requires an average of 3.72 s to complete a single decision-making inference. Such temporal inefficiency in reasoning fundamentally compromises the real-time performance requirements of autonomous vehicle decision systems, thereby significantly constraining the practical deployment of LLM experts. In contrast, the proposed method demonstrates an average decision-making inference time of merely 0.01 s per time step. These results show that the post-trained LGDRL not only exhibits superior testing performance among post-trained DRL agents but also demonstrates remarkable computational efficiency, substantially outperforming the LLM expert in terms of real-time decision-making capabilities.

\begin{figure}[t]
\centering
\includegraphics[width=0.97\linewidth]{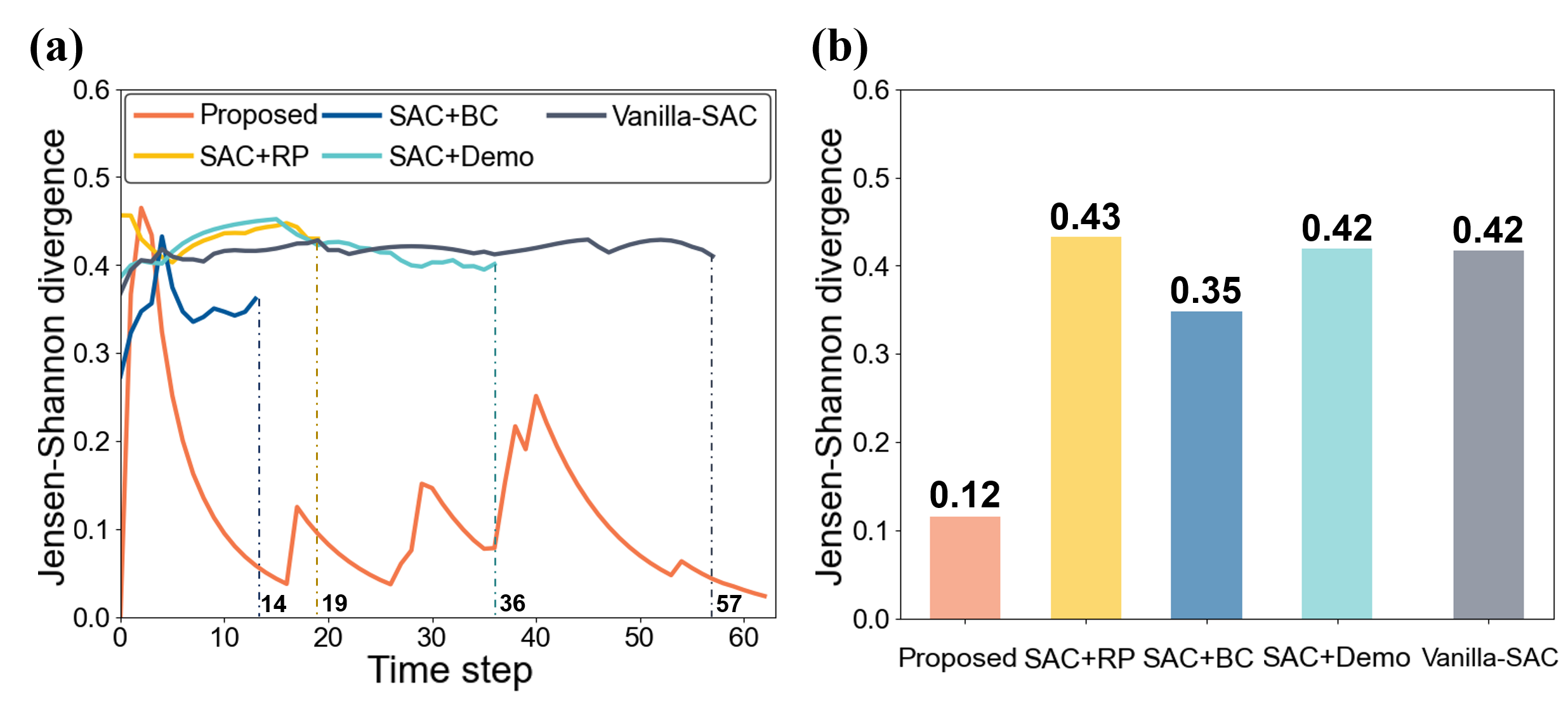}
\caption{The policy differences between the post-trained DRL agents and the LLM expert. (a) The curves of the JS divergence between the post-trained DRL agents and the LLM expert over time steps. The proposed method successfully completes the testing episode. The SAC+BC, SAC+RP, SAC+Demo and Vanilla-SAC fail at time step 14, 19, 36 and 57, respectively. (b) Average JS divergences of all time steps.}
\label{fig:moment}
\end{figure}

To validate the proposed LGDRL method's capability of effectively utilizing expert guidance for learning a decision-making policy similar to the LLM expert, all post-trained DRL agents and the LLM expert are independently tested for one episode within an identical driving scenario. The policy differences between these post-trained DRL agents and the LLM expert are then quantified and compared in this testing episode. The JS divergence, calculated by (\ref{eq:constrain_2}), is employed to quantitatively measure the policy differences. The results are illustrated in Fig. \ref{fig:moment}. As depicted in Fig. \ref{fig:moment}(a), only the proposed method successfully completes the entire testing episode, whereas the SAC+BC, SAC+RP, SAC+Demo, and Vanilla-SAC methods terminate early at time steps 14, 19, 36, and 57, respectively. The JS divergence of the proposed method is significantly smaller compared to other DRL baseline methods in the testing episode. Fig. \ref{fig:moment}(b) presents the average JS divergence across all time steps, calculated by normalizing the cumulative JS divergence by the total number of time steps. Notably, the proposed method exhibited the lowest JS divergence of 0.12, representing remarkable reductions of 32\%, 23\%, 30\%, and 30\% compared to SAC+RP, SAC+BC, SAC+Demo, and Vanilla-SAC, respectively. These results suggest that, compared to other methods, the proposed method more efficiently leverages expert guidance for policy learning, thereby deriving a decision-making policy that most closely aligned with the LLM expert.

\begin{figure}[t]
    \centering
    \captionsetup{justification=justified, font={small,rm}} 
    \subfloat[The normalized Q-value distribution of the proposed method at each time step within the testing episode. The actions of left lane change, decelerate, idle, accelerate, and right lane change are abbreviated as L, D, I, A, and R, respectively.]{\includegraphics[width=0.99\linewidth]{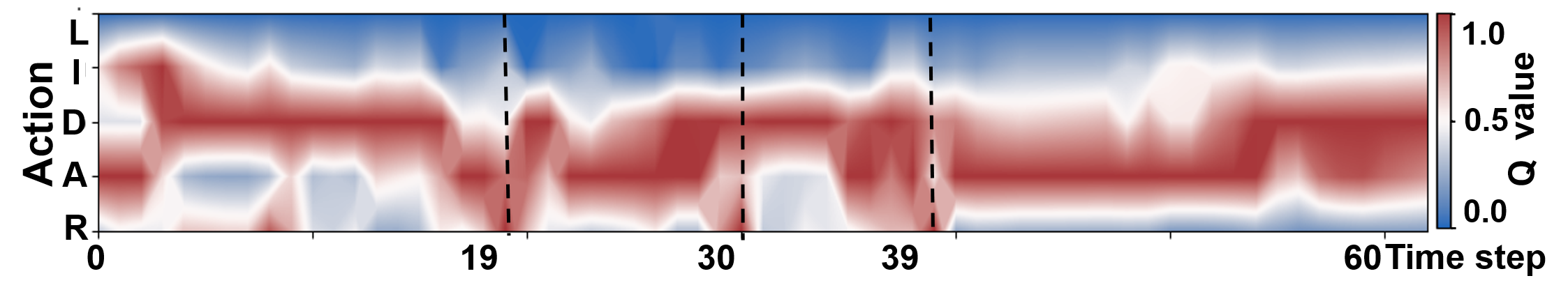}}
    \vspace{.04in}
    \subfloat[The action score distribution of the LLM expert at each time step within the testing episode. The action deemed most appropriate by the LLM expert at the current time step is assigned a score of 1, while the others are assigned a score of 0.]{\includegraphics[width=0.99\linewidth]{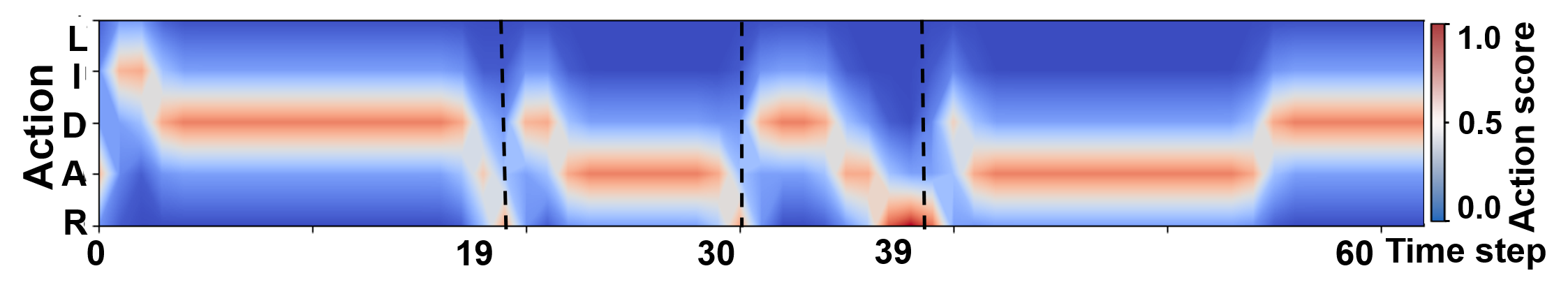}}
    \vspace{.04in}
    \subfloat[The driving scenarios at time step 19, 30, and 39 in the proposed method's testing episode.]{\includegraphics[width=0.99\linewidth]{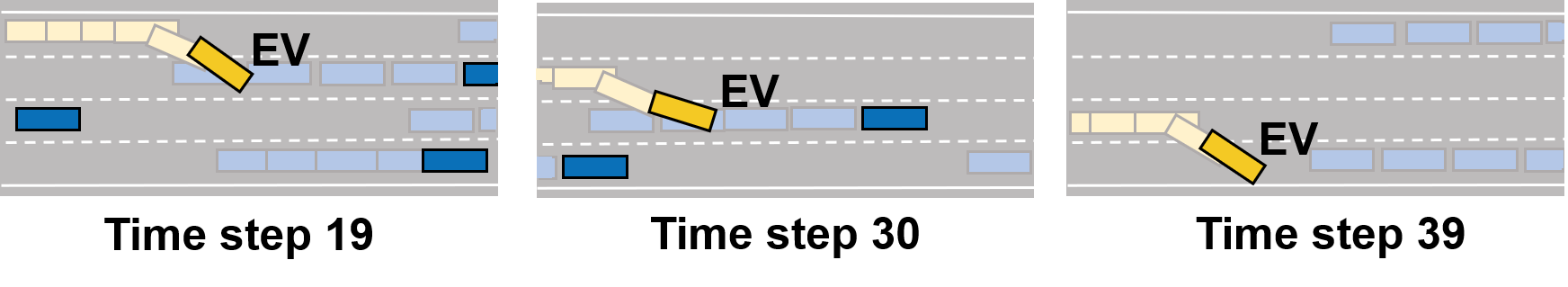}}
    \caption{Comparision of the decisions generated by the proposed method and the LLM expert.}
    \label{fig:success}
\end{figure}

Furthermore, the decisions generated by the proposed method and the LLM expert across each time step in the testing episode are investigated in detail. Fig. \ref{fig:success}(a) and Fig. \ref{fig:success}(b) show the normalized Q-value distribution of the proposed method and the action score distribution of the LLM expert in the testing episode, respectively. The proposed method relies on the Q-value to make decisions, selecting the action with the highest Q-value at each time step. The LLM expert evaluates the actions at each time step in the testing episode and the action deemed most appropriate by the LLM expert at the current time step is assigned a score of 1, while the others are assigned a score of 0. From an overall perspective, left lane change action is the most undesirable, this is because the target point is on the rightmost driving lane. By comparing the normalized Q-value distribution of the proposed method in Fig. \ref{fig:success}(a) with the action score distribution of the LLM expert in Fig. \ref{fig:success}(b), it is evident that, at each time step during the test episode, the proposed method consistently selects the actions which are deemed optimal by the LLM expert. Specifically, at time steps 19, 30, and 39, the proposed method chooses the right lane change action, which the LLM expert also identifies as the most appropriate action at these specific time steps. At a detailed level, Fig. \ref{fig:success}(c) illustrates these three key time steps during the testing episode of the proposed method. At time steps 19, 30, and 39, the right lane adjacent to the EV is relatively clear, providing the EV with an opportunity to change lanes to the right and move closer to the target point. Consequently, the proposed LGDRL method selects the right lane change action.

From these results, we can conclude that, compared to other DRL methods, our proposed LGDRL method not only achieves optimal testing performance in terms of task success rate and driving safety but also exhibits a decision-making policy that is most similar to the LLM expert, which demonstrates the proposed method's efficient utilization of the LLM expert's guidance in the policy learning process.

\begin{figure}[t]
\centering
\includegraphics[width=0.99\linewidth]{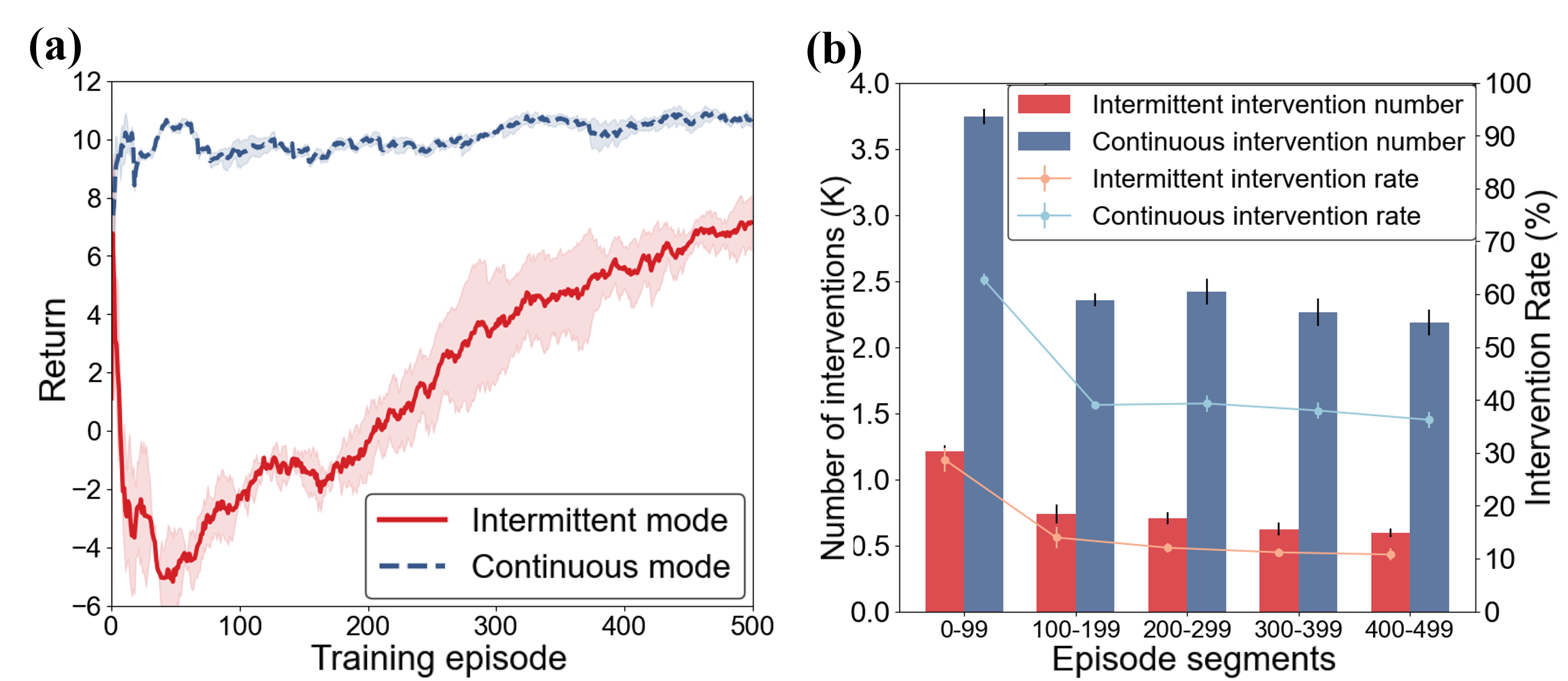}
\caption{The impact of different intervention modes on the training performance. (a) The curves of the episode return; (b) The intervention number and intervention rate in different training phases.}
\label{fig:diff_guidetype}
\end{figure}

\subsection{The Effects of Different Intervention Modes}

In Section \Romannum{4}-D, we introduce the expert-intervened interaction mechanism which enables the LLM expert to intervene in the interaction between the DRL agent and the environment during training in an intermittent mode. Here, we investigate the impact of different intervention modes on the training and testing performance of the proposed LGDRL method. In addition to the intermittent mode introduced in Section \Romannum{4}-D, we implement another intervention mode, named the continuous mode \cite{guide}. Unlike the intermittent mode which permits the LLM expert to intervene in only selected episodes during training, the continuous mode allows the LLM expert to intervene in any episode throughout the training process. Although this continuous mode facilitates the DRL agent sampling more successful data, it may potentially reduce its self-exploratory capability.

The proposed method is trained with the same LLM expert but different intervention modes: intermittent mode and continuous mode. The metrics used to evaluate the training performance include return, intervention number, and intervention rate. The training results are depicted in Fig. \ref{fig:diff_guidetype}. As shown in Fig. \ref{fig:diff_guidetype}(a), the return curves exhibit distinct patterns. Specifically, continuous mode results in a more stable and higher return compared to intermittent mode. The return of continuous mode remains consistently around 10 throughout the training process. In contrast, the return of intermittent mode fluctuates significantly during the first 50 episodes and eventually settles around 6. The superior performance of continuous mode is attributed to the more consistent and immediate interventions provided by the LLM expert, which help the DRL agent more effectively avoid suboptimal actions. On the other hand, intermittent mode has episodes without expert intervention, leading to more frequent suboptimal decisions and slower learning progress. The number and rate of intervention in different training phases are further investigated, as shown in Fig. \ref{fig:diff_guidetype}(b). The number and rate of intervention for both modes are significantly higher during the initial training phase (episodes 0-99) compared to subsequent phases. This trend is due to the DRL agent engaging in extensive exploration during the early stages of training. Specifically, the intervention number of continuous mode decreases by approximately 40.54\% from the initial phase to the final phase (episodes 400-499) and the intervention rate declines from about 62\% to 38\%. In contrast, the intermittent mode exhibits lower intervention rate and intervention number throughout the training period. The intervention number of intermittent mode shows a reduction of about 50\% and the intervention rate drops from approximately 30\% to about 10\%. These results imply that the training performance of the proposed method improves progressively under both intervention modes. However, the continuous mode exhibits more frequent expert interventions.
\begin{table}[t]
\caption{Testing Performance of the Proposed Method Trained Under Different Intervention Mode\label{tab:test_diffguidetype}}
\centering
\begin{tabular}{cccc}
\hline
    &\makecell{Success\\Rate (\%)} &\makecell{Collision\\Rate (\%)}  & \makecell{Return\\}  
    \\ \hline
Intermittent Mode    & \textbf{90}         & \textbf{10}             & \textbf{6.87±8.27}    \\
Continuous Mode & 44                  & 56                      & 0.75±10.57            \\ \hline
\label{table:diff_guidetype_test}
\end{tabular}
\end{table}

Subsequently, the post-trained LGDRL agents trained under the two intervention modes are tested 50 episodes. Success rate, collision rate, and return are selected as the evaluation metrics for the testing performance. The results are listed in Table \ref{table:diff_guidetype_test}. The continuous mode underperforms the intermittent mode across all metrics, which demonstrates a significant deviation from the performance observed in the training stage. This discrepancy can be explained by the excessive interventions inherent in the continuous mode, which hinder the agent to fully explore the environment during training, thereby reducing its testing performance. 

These results demonstrate that the intermittent intervention mode used in our proposed expert-intervened interaction mechanism can improve the training performance of DRL through an appropriate number of interventions, while also preserving DRL agent's exploratory capability. This ensures that in the testing stage, without expert assistance, DRL agent can still maintain the same good performance as in the training stage.

\subsection{Ablation Study}
\begin{figure}[t]
\centering
\includegraphics[width=0.99\linewidth]{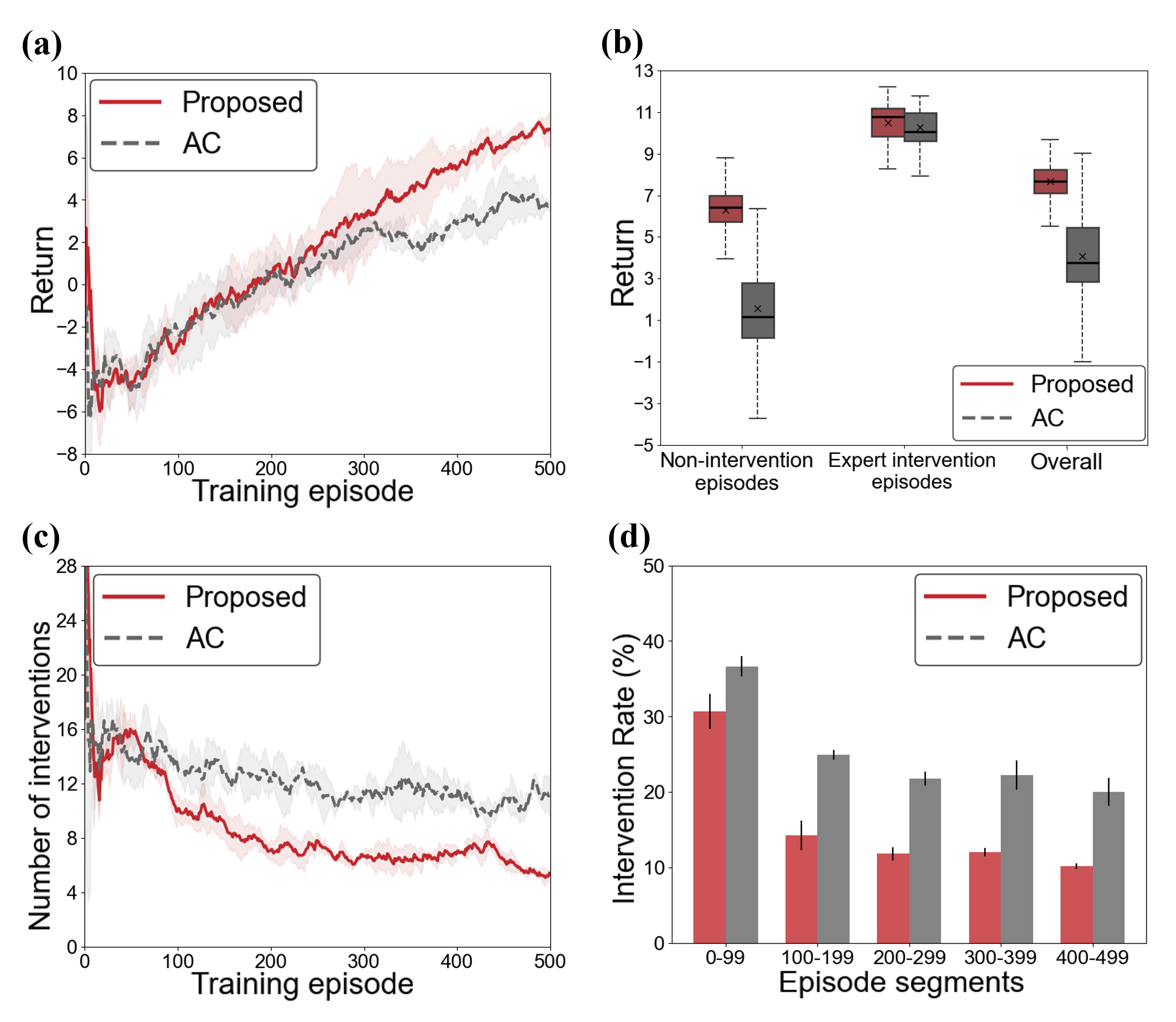}
\caption{The training performance of the LGDRL and AC methods. (a) The curves of the episode return; (b) Box plot of episode returns, categorized by non-intervention episodes, expert intervention episodes, and all episodes combined; (c) The curves of intervention number per episode; (d) Bar plot of the intervention rates at different stages of training.}
\label{fig:ablation}
\end{figure}

An ablation study is performed to further investigate the impact of expert policy constraint on the training performance of our proposed LGDRL method. Based on the LGDRL method, an AC method that excludes the expert constraint component is constructed for comparison. Both methods employ the same LLM expert and the same intermittent intervention mode in the training stage. The training performance of LGDRL and AC methods is shown in Fig. \ref{fig:ablation}. 

As shown in Fig. \ref{fig:ablation}(a), the returns of the two methods are similar in the early stages of training. However, as training progresses, the proposed method significantly outperforms the AC method. The returns obtained in episodes with and without expert intervention are illustrated in Fig \ref{fig:ablation}(b). The average return in episodes with expert intervention is close. However, in episodes without expert intervention, the average return of the proposed method is approximately 4.5 higher than that of the AC method. Considering the overall returns, the proposed method achieves an average return approximately 3.8 higher than the AC method. Therefore, the disparity in episodes without expert intervention is the primary reason for the overall performance difference between the two methods. This indicates that the expert policy constraint component is crucial for effectively leveraging expert prior knowledge to enhance training performance. Without this component, the AC method struggles to achieve optimal performance in the absence of expert intervention. The intervention number and intervention rate in the training stage for both methods are shown in Fig. \ref{fig:ablation}(c) and Fig. \ref{fig:ablation}(d), respectively. The proposed method exhibits significantly lower intervention number and intervention rate compared to the AC method. These results demonstrate that the expert policy constraint component significantly improves the DRL agent's efficiency in utilizing expert guidance, enabling the DRL agent to learn a policy similar to the expert, thereby gradually reducing its reliance on expert guidance.

\section{Conclusion}
In this article, we propose a novel LGDRL framework to address the lane change decision-making problem of autonomous vehicles. Within this framework, we develop the LLM-based driving expert to provide guidance for the learning process of DRL. To more efficiently leverage the guidance from the LLM expert, we introduce the innovative expert policy constrained actor-critic algorithm. Moreover, we design the expert-intervened interaction mechanism which allows the LLM expert to intermittently intervene in the DRL's interaction with the environment, preventing the DRL agent from becoming trapped in local optima. Comparative experiments with state-of-the-art methods demonstrate that our proposed method not only achieves optimal training and testing performance but also exhibits superior learning efficiency. We also find that the intermittent intervention mode enables the DRL agent to maintain its self-exploration capabilities, enabling the DRL agent to perform well even in the absence of the LLM expert. The ablation study indicates that the policy constraint component used in our method significantly enhances the DRL agent's ability to utilize expert guidance. Future work will focus on exploring the applicability of this framework to other complex driving scenarios.

\bibliographystyle{IEEEtran}
\bibliography{IEEEabrv,IEEEexample}

\end{document}